\documentclass[11pt]{article}

\usepackage[preprint]{acl}

\usepackage{times}
\usepackage{latexsym}
\usepackage{subcaption}
\usepackage{multirow}
\usepackage{booktabs}
\usepackage{fontawesome5}

\usepackage[T1]{fontenc}

\usepackage[utf8]{inputenc}

\usepackage{microtype}

\usepackage{inconsolata}

\usepackage{graphicx}
\usepackage{enumitem}
\usepackage{amsmath}
\usepackage{amssymb}
\usepackage{tabularx}
\usepackage{soul}

\title{\textit{Do LLMs Really Know What They Don't Know?} \\Internal States Mainly Reflect Knowledge Recall Rather Than Truthfulness}

\author{Chi Seng Cheang$^1$~~~
        Hou Pong Chan$^2$~~~
        Wenxuan Zhang$^3$~~~
        Yang Deng$^1$\thanks{Corresponding author.}~~~ \\
  $^1$Singapore Management University \quad    
  $^2$DAMO Academy, Alibaba Group \\
    $^3$Singapore University of Technology and Design  \\
    \texttt{cs.cheang.2025@phdcs.smu.edu.sg, houpong.chan@alibaba-inc.com} \\
          \texttt{wxzhang@sutd.edu.sg, ydeng@smu.edu.sg } }

\begin{document}
\maketitle
\begin{abstract}

Recent work suggests that LLMs ``know what they don't know'', positing that hallucinated and factually correct outputs arise from distinct internal processes and can therefore be distinguished using internal signals.
However, hallucinations have multifaceted causes: beyond simple knowledge gaps, they can emerge from training incentives that encourage models to exploit statistical shortcuts or spurious associations learned during pretraining.
In this paper, we argue that when LLMs rely on such learned associations to produce hallucinations, their internal processes are mechanistically similar to those of factual recall, as both stem from strong statistical correlations encoded in the model's parameters.
To verify this, we propose a novel taxonomy categorizing hallucinations into Unassociated Hallucinations (UHs), where outputs lack parametric grounding, and Associated Hallucinations (AHs), which are driven by spurious associations. Through mechanistic analysis, we compare their computational processes and hidden-state geometries with factually correct outputs.
Our results show that hidden states primarily reflect whether the model is \textit{recalling parametric knowledge} rather than the truthfulness of the output itself. Consequently, AHs exhibit hidden-state geometries that largely overlap with factual outputs, rendering standard detection methods ineffective. In contrast, UHs exhibit distinctive, clustered representations that facilitate reliable detection.

\texttt{\faGithub \  \url{https://github.com/AndyCheang/knowledge-recall-vs-truthfulness}}

\end{abstract}


\section{Introduction}

\begin{figure}[t]
\setlength{\abovecaptionskip}{5pt}   
\setlength{\belowcaptionskip}{0pt}
\centering
    \includegraphics[width=1.05\linewidth]{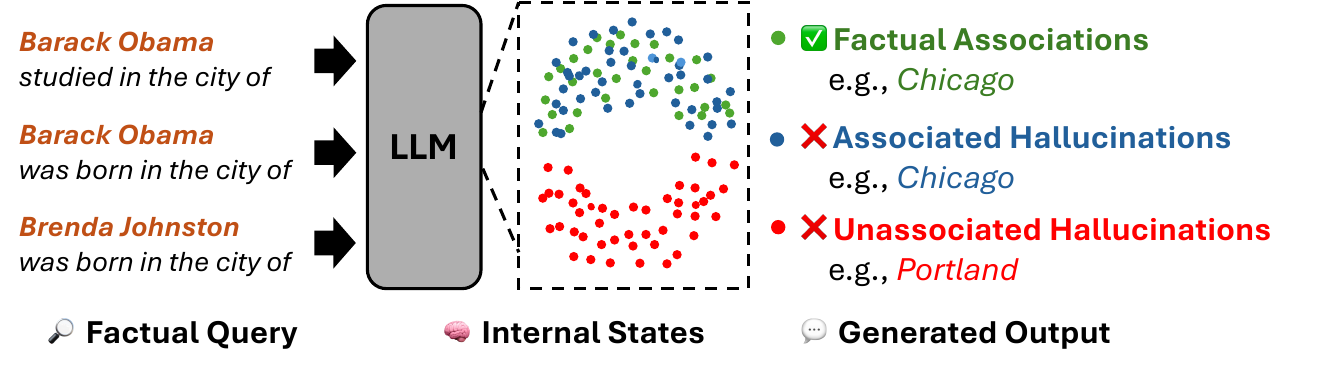}
\caption{
Illustration of three categories of knowledge. Associated hallucinations follow similar internal knowledge recall processes with factual associations, while unassociated hallucinations arise when the model’s output is detached from the input.
}
\label{fig:intro}
\vspace{-3mm}
\end{figure}

Large language models (LLMs) are notorious for producing hallucinations \cite{hallu-survey1,hallu-survey2}, where generated outputs appear plausible but are factually incorrect.
Recent research suggests that LLMs' internal states contain signals correlated with factual correctness, enabling hallucination detection using internal representations, such as residual streams \cite{azaria2023internal,gottesman2024estimatingknowledgelargelanguage}, attention weights \cite{yuksekgonul2023attention}, and output token logits \cite{orgad2024llms,DBLP:journals/corr/abs-2307-03987}.
However, since LLMs are not explicitly trained to represent truthfulness, it remains unclear whether these signals genuinely reflect truthfulness or instead capture other confounding factors. 
Understanding what these signals actually encode is critical for reliably deploying LLMs in real-world applications.

In this work, we argue that such internal signals primarily reflect the model's internal process of recalling parametric knowledge, rather than truthfulness itself.
Consequently, these signals can reliably detect hallucinations only when hallucinated and factually correct outputs are produced by distinct internal mechanisms.
For example, as shown in Figure~\ref{fig:intro}, given the prompt \textit{``Brenda Johnston was born in the city of''}, a model that lacks the relevant factual knowledge about the subject (\textit{``Brenda Johnston''}) may hallucinate a completion such as \textit{``Portland''}.
In contrast, given the prompt \textit{``Barack Obama studied in the city of''}, the model can leverage knowledge encoded about the subject (\textit{``Barack Obama''}) to produce the factually correct output \textit{``Chicago''}.
These two cases are likely supported by different internal mechanisms: the former lacks knowledge about the subject entity, while the latter relies on encoded knowledge relevant to the queried subject. As a result, internal representations can reflect this difference in how the model processes the subject entity, enabling these cases to be distinguished. 

However, hallucinations do not always arise from missing knowledge.
When a model exploits learned statistical shortcuts or spurious correlations \citep{lin2021truthfulqa,kang2023impact,DBLP:conf/emnlp/CheangCW0LS0C23}, the resulting hallucinations may be produced through mechanisms similar to those underlying factual recall.
For instance, \textit{``Barack Obama''} frequently co-occurs with \textit{``Chicago''} in the model's pre-training corpora. The model can leverage this statistical association to produce a factually correct output (e.g., \textit{``Barack Obama studied in the city of Chicago.''}), but it may also leverage the same association to produce a hallucinated response (e.g., \textit{``Barack Obama was born in the city of Chicago.''}).
In both cases, the model relies on the same encoded statistical association involving the subject entity.
Consequently, the resulting internal representations may not provide reliable signals to distinguish hallucinated outputs from factual ones, limiting the effectiveness of existing representation-based hallucination detection methods.

Based on this observation, we hypothesize that the effectiveness of representation-based hallucination detection depends on how the model leverages its parametric knowledge when producing a response, particularly whether the generated output is driven by learned associations involving the subject entity. To investigate this hypothesis, we go beyond labeling outputs solely by factual correctness and instead categorize them according to their relationship with the subject entity through a causal intervention. 
Specifically, we label factually correct outputs as \textbf{Factual Associations (FAs)}. For outputs that are factually incorrect, we further classify them as \textbf{Unassociated Hallucinations (UHs)}, whose outputs lack strong learned associations with the subject entity, and \textbf{Associated Hallucinations (AHs)}, which are driven by strong but spurious associations involving the subject entity.

Using this taxonomy, we conduct mechanistic and empirical analyses of these knowledge categories, yielding three key observations:
First, \textbf{AHs and FAs share highly similar internal processes and representational geometries.} Building on the analytical framework of \citet{DBLP:conf/emnlp/GevaBFG23}, we examine the internal mechanisms underlying model predictions by tracing how information propagates across layers and token positions during inference. We observe that because AHs and FAs are both driven by learned associations with the subject, their hidden state representations overlap in the hidden space. In contrast, UHs lack this reliance on subject associations and are instead generated through a different internal process, allowing them to remain more separable in the representation space.

Second, \textbf{existing hallucination detection methods struggle to distinguish AHs from FAs.} Since these methods rely on internal states that primarily reflect the process of knowledge recall rather than truthfulness, their performance degrades significantly for AH samples (AUROC $\approx 0.48$--$0.69$ for LLaMA). However, UHs are more reliably detected (AUROC $\approx 0.86$--$0.93$) due to their more separable representational geometry.

Third, \textbf{representational overlap constrains the effectiveness of refusal tuning for AHs.} We compare tuning effectiveness under two settings: (i) training the model to refuse AHs, and (ii) training the model to refuse UHs. In both settings, the model is trained to maintain its original factual responses for FAs. Because UH representations are more separable from FAs, the model can successfully learn distinct generative behaviors, achieving an 82\% refusal rate on UH samples. Conversely, because AH representations overlap substantially with FAs, the model struggles to differentiate them to learn refusal behaviors, resulting in a refusal rate of only 33\% for AH samples.

\section{Related Work}

Existing hallucination detection methods can be broadly categorized into two types: \textit{representation-based} and \textit{confidence-based}. 
\textbf{Representation-based methods} assume that an LLM's internal hidden states can reflect the correctness of its generated responses. These approaches train a classifier (often a linear probe) using the hidden states from a set of labeled correct/incorrect responses to predict whether a new response is hallucinatory~\cite{li2023inference,azaria2023internal,DBLP:conf/acl/SuWAH00024,ji-etal-2024-llm,DBLP:conf/iclr/0026L0GWTFY24,DBLP:conf/acl/NiBGYBC25,xiao2025analyzing}.
\textbf{Confidence-based methods}, in contrast, assume that a lower confidence during the generation led to a higher probability of hallucination. 
These methods quantify uncertainty through various signals, including: (i) token-level output probabilities~\citep{DBLP:conf/eacl/GuerreiroVM23, DBLP:journals/corr/abs-2307-03987, orgad2024llms}; (ii) directly querying the LLM to verbalize its own confidence~\citep{DBLP:journals/tmlr/LinHE22, DBLP:conf/emnlp/TianMZSRYFM23, DBLP:conf/iclr/XiongHLLFHH24, DBLP:conf/nips/0004CQN024, DBLP:conf/acl/NiBGC24,DBLP:conf/naacl/ZhaoY0XMWCRY24}; or (iii) measuring the semantic consistency across multiple outputs sampled from the same prompt~\citep{DBLP:conf/emnlp/ManakulLG23, DBLP:conf/iclr/KuhnGF23, DBLP:journals/corr/abs-2311-01740,DBLP:journals/corr/abs-2402-10612}. A response is typically flagged as a hallucination if its associated confidence metric falls below a predetermined threshold.

However, a growing body of work reveals a critical limitation: even state-of-the-art LLMs are poorly calibrated, meaning their expressed confidence often fails to align with the factual accuracy of their generations~\citep{DBLP:conf/nips/KapoorGRCPBWDGW24, DBLP:conf/iclr/XiongHLLFHH24, DBLP:conf/emnlp/TianMZSRYFM23}. This miscalibration limits the effectiveness of confidence-based detectors and raises a fundamental question about the extent of LLMs' self-awareness of their knowledge boundary, \textit{i.e.}, whether they can reliably ``\textit{know what they don't know}'' \cite{DBLP:conf/acl/YinSGWQH23,DBLP:conf/acl/LiZZLXNCD25}. Despite recognizing this problem, prior work does not provide a mechanistic explanation for its occurrence. To this end, our work addresses this explanatory gap by employing mechanistic interpretability techniques to trace the internal computations underlying knowledge recall within LLMs.

\section{Dataset Construction}
\label{sec:data}

In this section, we outline our dataset construction for mechanistic and empirical analyses under two conditions: hallucinations produced with and without leveraging the learned associations related to the subject entity.
Given an input query $q$, the ground-truth answer $y$, and the model's response $\hat{y}$, standard evaluation of hallucination detection methods typically assigns a factual correctness label by comparing $\hat{y}$ with $y$.
To study hallucinations produced through different internal mechanisms, we go beyond factual correctness: for each hallucinatory sample, we perform a causal intervention to estimate its reliance on learned subject associations and categorize it accordingly.

\subsection{Data Collection}

\paragraph{Factual Query Prompt Creation}
We focus on a knowledge-based question answering setting, where each example corresponds to a knowledge triple \texttt{(subject, relation, object)} $(s, r, o)$. 
To construct factual query prompts, we first collect knowledge triples from Wikidata \cite{DBLP:journals/cacm/VrandecicK14}. Each $(s,r)$ pair is then converted into a cloze-style factual query $q$ using a handcrafted prompt template for each relation $r$. The corresponding object $o$ is treated as the ground-truth answer $y$.
To ensure a well-defined evaluation setting, we follow \citet{DBLP:journals/corr/abs-2503-15299} and select only relations for which the correct object is objectively verifiable.
Details on relation selection and prompt templates are provided in Appendix~\ref{sec:relation-and-prompt-template}.

\paragraph{Generating Model Responses}
For each query, we prompt LLMs to generate a response $\hat{y}$ using greedy decoding.
We conduct experiments on two widely adopted open-source LLMs: LLaMA-3 \cite{DBLP:journals/corr/abs-2407-21783} and Mistral-v0.3 \cite{jiang2023mistral7b}.
Due to space constraints, full implementation details are provided in Appendix~\ref{app:implement}.

\subsection{Categorization of Knowledge}
\label{sec:categorization}

We categorize each response based on two criteria: (1) \textit{factual correctness} and (2) \textit{reliance on subject representations}. 
Each sample is then categorized into one of the following categories:

\begin{itemize}[leftmargin=*,nosep]

    \item \textbf{Factual Associations (FA)} refer to factual knowledge that is reliably stored in the parameters or internal states of an LLM and can be recalled to produce correct, verifiable outputs.

    \item \textbf{Associated Hallucinations (AH)} refer to non-factual content produced when an LLM relies on input-triggered parametric associations. 

    \item \textbf{Unassociated Hallucinations (UH)} refer to non-factual content produced without reliance on parametric associations to the input.

\end{itemize}

\begin{figure*}[t]
\setlength{\abovecaptionskip}{5pt}   
\setlength{\belowcaptionskip}{0pt}
  \centering
  \begin{subfigure}[t]{0.3\linewidth}
    \includegraphics[width=\linewidth]{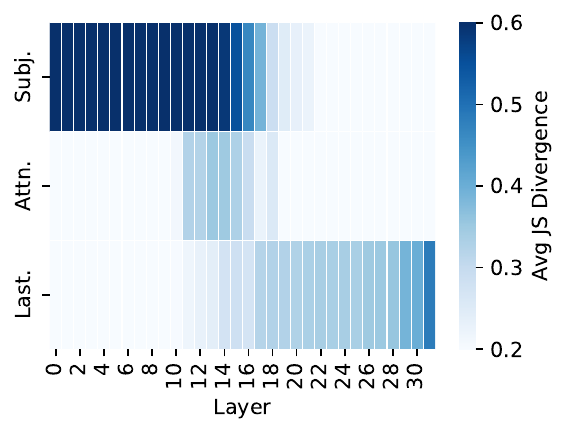}
    \caption{Factual Associations}
    \label{fig:interventions-llama-correct}
  \end{subfigure} \hfill
  \begin{subfigure}[t]{0.3\linewidth}
    \includegraphics[width=\linewidth]{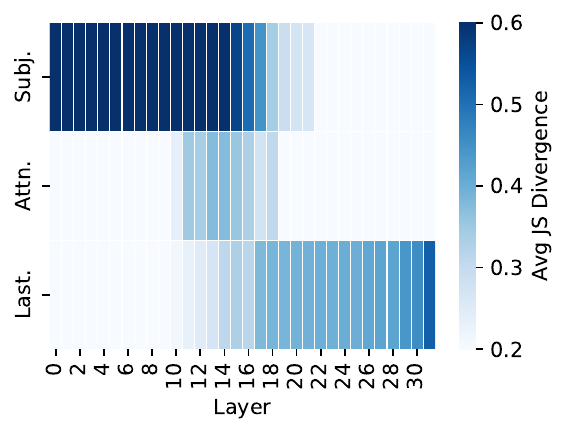}
    \caption{Associated Hallucinations}
    \label{fig:interventions-llama-ah}
  \end{subfigure} \hfill
  \begin{subfigure}[t]{0.3\linewidth}
    \includegraphics[width=\linewidth]{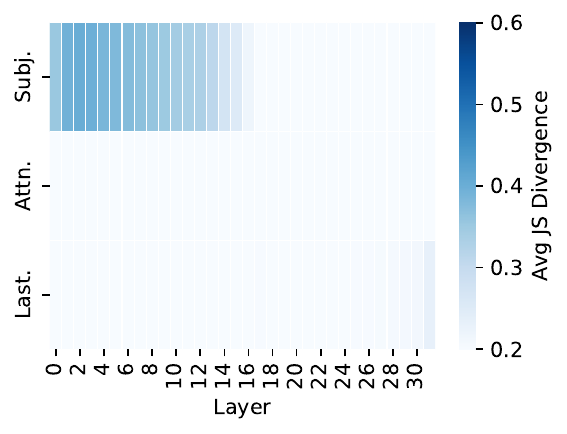}
    \caption{Unassociated Hallucinations}
    \label{fig:interventions-llama-uh}
  \end{subfigure} 
  \caption{Effect of interventions across layers of LLaMA-3-8B. The heatmap shows JS divergence between the  output distribution before and after intervention. Darker color indicates that the intervened hidden states are more causally influential on the model’s predictions.
Top row: patching representations of subject tokens. 
Middle row: blocking attention flow from subject to the last token.  
Bottom row: patching representations of the last token.}
  \label{fig:interventions}
\end{figure*}

\subsection{Labeling Procedure}\label{sec:labeling}
We detail the labeling procedure as follows:

\paragraph{Factual Correctness}
 
We assess the factual correctness of model responses following the framework proposed by \citet{DBLP:journals/corr/abs-2503-15299} (details in Appendix \ref{app:label}). If a model's response $\hat{y}$ matches the ground-truth answer $y$, it is labeled as a \textbf{Factual Association (FA)}. Otherwise, the prediction is considered a hallucination and is further categorized based on its reliance on subject representations.

\paragraph{Subject Representation Reliance}

\citet{DBLP:conf/emnlp/GevaBFG23} analyzes LLMs' internal activations to characterize mechanisms underlying factual generation.
Given a factual query prompt (e.g., \textit{``Barack Obama was born in the city of''}), a model typically follows three steps to produce the correct completion (e.g., \textit{``Honolulu''}).
First, early layers of LLMs construct enriched subject representations encoding attributes specific to the subject (e.g., \textit{``Barack Obama''}).
Second, attention modules in middle layers propagate this subject information to the prediction position (i.e., the last token).
Finally, upper layers extract the relevant attribute (e.g., \textit{``Honolulu''}) from the propagated information to produce the next token.

\begin{table}[t]
\setlength{\abovecaptionskip}{5pt}   
\setlength{\belowcaptionskip}{0pt}
\centering
\resizebox{\columnwidth}{!}{%
\begin{tabular}{@{}lrr@{}}
\toprule
 & \textbf{LLaMA-3-8B} & \textbf{Mistral-7B-v0.3} \\
\midrule
Factual Association      & 3,506  & 3,354  \\
Associated Hallucination   & 1,406  & 1,284  \\
Unassociated Hallucination & 7,381 & 7,655 \\
\midrule
\textbf{Total}             & 12,293 & 12,293 \\
\bottomrule
\end{tabular}%
}
\caption{Dataset statistics across categories.}
\label{tab:dataset-statistics}
\vspace{-3mm}
\end{table}

While this mechanism has been established for factual generation, the internal processes underlying hallucinations remain less understood. We hypothesize that hidden states primarily reflect \textit{how} the model leverages its parametric knowledge, rather than whether the output is factually correct. Consequently, hallucinations produced through statistical shortcuts or spurious correlations may follow internal processes similar to factual recall.

To test this hypothesis, we categorize hallucinatory outputs based on their reliance on learned associations with the subject entity. Specifically, we perform a causal intervention by blocking attention from the subject tokens and measuring the resulting shift in the output distribution. Because attention is the primary mechanism through which subject representations interact with subsequent tokens, a large distributional shift under masking indicates a strong reliance on the subject representation. We quantify this shift using the Jensen--Shannon (JS) divergence between the original and masked output distributions. 
We set the threshold to the average JS divergence observed for all correct answers (FAs): hallucinations are categorized as \textbf{Associated Hallucinations (AH)} if the JS divergence exceeds this threshold; otherwise, they are categorized as \textbf{Unassociated Hallucinations (UH)}.
Table \ref{tab:dataset-statistics} summarizes the final data statistics.

\section{Analysis of Internal States in LLMs}
\label{sec:analyzing-internal-states}

As established by \citet{DBLP:conf/emnlp/GevaBFG23}, factual generation relies on three components: (i) the enrichment of subject representations in the early layers, (ii) an attention flow from the subject to the last token in the middle layers, and (iii) the decoding of the last token representation in the upper layers. In this section, we first visualize this overall information flow across all categories (\S \ref{sec:localizing-the-information-flow}), and analyze each component in \S \ref{sec:subject-analysis} through \S \ref{sec:last-token-representation}. Parallel experimental results on Mistral are summarized in Appendix \ref{app:mistral}.

\subsection{Visualization of Information Flow}
\label{sec:localizing-the-information-flow}

In \S \ref{sec:labeling}, we categorized hallucinations by intervening on the attention flow from the subject tokens. 
To provide a more granular understanding of the internal mechanisms established by \citet{DBLP:conf/emnlp/GevaBFG23}, we perform fine-grained causal interventions on all three components across each knowledge category. This allows us to examine whether the enrichment and decoding phases remain consistent between Associated Hallucinations (AHs) and Factual Associations (FAs).

\paragraph{Experimental Setup}
We measure the importance of each component by observing the resulting shift in the output distribution after intervention; a large shift indicates the component is critical for the prediction, while a small shift suggests a limited role.
We quantify each shift using Jensen-Shannon (JS) divergence between the original and intervened output distributions. Specifically, for subject-token and last-token representations, we corrupt the corresponding hidden states at each layer $\ell$. For the attention mechanism, we follow \citet{DBLP:conf/emnlp/GevaBFG23} to mask the attention flow between the subject and last tokens at layer $\ell$, using a window size of 5 layers.

\paragraph{Experimental Results}
As shown in Figure~\ref{fig:interventions}, FAs and AHs share highly similar information flows.  
The components that are crucial for producing factually correct outputs are also crucial for producing AHs. This pattern suggests that, despite being factually incorrect, AHs are generated through the same underlying internal mechanisms as factual recall. In contrast, UHs display substantially weaker information flow, reflecting their limited reliance on the subject entity during generation. 

These findings indicate that hidden states primarily capture \textbf{whether the model is recalling parametric knowledge}, rather than \textbf{whether the generated output is factually correct}. 
These results challenge the prevailing assumption in prior works \cite{azaria2023internal, gottesman2024estimatingknowledgelargelanguage, yuksekgonul2023attention, orgad2024llms, varshney2023stitch} that internal representations inherently encode factual correctness. Instead, our results show that factual outputs and association-driven hallucinations can share nearly identical internal computational pathways.

It is worth noting that the strong causal effect of attention-flow interventions for FAs and AHs is expected, since our labeling procedure already uses attention masking to distinguish AHs from UHs.
However, interventions on subject-token representations and last-token hidden states provide an independent validation of our hypothesis.
These components remain highly influential for both FAs and AHs, confirming that the shared mechanism extends beyond the attention pathways used during labeling.

\subsection{Analysis of Subject Representations}
\label{sec:subject-analysis}

Having established the overall information flow, we next analyze the subject representations in the early layers. Our goal is to understand why the internal behavior at subject positions is highly similar for AHs and FAs, yet differs substantially for UHs. 
To this end, we first examine the norm of these representations (\S \ref{sec:subject-representation-norm-analysis}), and then analyze how this behavior correlates with subject prevalence in the pre-training data (\S \ref{sec:correlation-with-subject-popularity}).

\subsubsection{Norm of Subject Representations}
\label{sec:subject-representation-norm-analysis}

We measure the average $L_2$ norm of subject representations for each knowledge category across layers (details in Appendix \ref{sec:appendix_norm_details}).
To facilitate direct comparison, we report the $L_2$ norms of both hallucination types (AHs and UHs) as ratios relative to the norm of factually correct predictions (FAs).
A ratio close to 1 indicates norms comparable to FAs, while ratios below 1 indicate smaller norms.

\begin{figure}[t]
\setlength{\abovecaptionskip}{0pt}   
\setlength{\belowcaptionskip}{0pt}
\centering
    \includegraphics[width=0.8\linewidth]{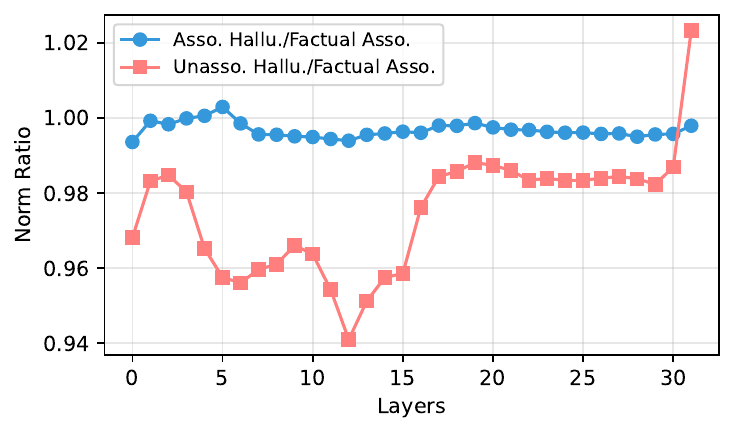}
\caption{
Norm ratio curves of subject representations in LLaMA-3-8B, comparing AHs and UHs against FAs as the baseline. 
}
\label{fig:subject_resp_norm_ratio}
\end{figure}

As shown in Figure~\ref{fig:subject_resp_norm_ratio}, for LLaMA-3-8B, the norm of AH samples is highly comparable to that of FAs, with ratios remaining close to 1 across layers.
In contrast, UH samples consistently exhibit smaller norms than FAs in nearly all layers, with the ratio dropping to approximately 0.95 in the middle layers (layers 10–15).
This suggests that \textbf{AHs and FAs activate subject representations with similar strength, whereas UHs involve weaker subject encoding}. 
We further analyze the subspace overlap between subject hidden states and MLP weight matrices in Appendix~\ref{app:subject-representation-subspace-overlap}, which provides additional evidence that AHs and FAs occupy highly similar representational regions.

\subsubsection{Correlation with Subject Popularity}
\label{sec:correlation-with-subject-popularity}

To explain why AHs exhibit subject representation norms comparable to FAs while UHs remain weaker, we hypothesize that this difference arises from subject popularity in the pre-training data.
Prior work \cite{kang2023deep} shows that entities appearing more frequently during pre-training tend to produce hidden representations with larger norms.
Accordingly, we expect that FAs and AHs predominantly involve high-popularity entities, while UHs are more common for low-popularity entities.

\begin{figure}[t]
\setlength{\abovecaptionskip}{5pt}   
\setlength{\belowcaptionskip}{0pt}
\centering
    \includegraphics[width=0.8\linewidth]{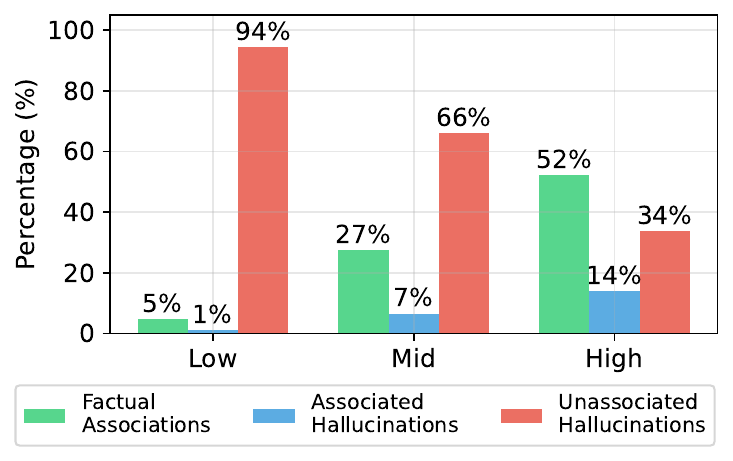}
\caption{
Sample distribution across different subject popularity (low, mid, high) in LLaMA-3-8B, measured by monthly Wikipedia page views. }
\label{fig:error-distribution}
\end{figure}

Following \citet{mallen2022not}, we use average monthly Wikipedia page views as a proxy for subject frequency during pre-training.
We partition subjects into three popularity bins and analyze the distribution of FAs, AHs, and UHs across them. The distribution in Figure~\ref{fig:error-distribution} confirms that the prevalence of both FAs and AHs is strongly associated with subject popularity. In the lowest popularity bin, both FAs and AHs are rarely observed (5\% for FAs and 1\% for AHs). In contrast, their prevalence significantly increases in the highest popularity bin (52\% for FAs and 14\% for AHs). However, the prevalence of UHs shows an inverse trend: while the lowest popularity bin is dominated by UHs (94\%), their prevalence drops to 34\% in the highest popularity bin.

This observation highlights a critical implication: if high-popularity subjects account for a substantial portion of real-world language model usage, AHs may constitute a substantial portion of hallucinations encountered in practical deployments. 
Meanwhile, as AHs arise mainly on popular subjects, they are often indistinguishable from FAs by popularity-based heuristics, contradicting prior work \cite{mallen2022not} that links popularity to hallucinations.

\subsection{Visualization of Attention Flow}
\label{sec:attention}

Having examined how the model forms subject representations, we next visualize how this information is propagated to the last token of the input where the model generates the object of a knowledge tuple. 
While the strong reliance of AHs on this attention flow is an expected consequence of our causal intervention labeling, visualizing this mechanism explicitly illustrates the divergence between AHs/FAs and UHs, providing a more comprehensive view of the internal mechanism behind how the model produces outputs for each category.

To quantify the specific contribution from subject tokens $(s_1, ..., s_n)$ to the last token, we compute the attention contribution from subject tokens to the last position:  
\begin{equation}\small
\mathbf{a}^\ell_{\text{last}} = \sum\nolimits_{k} \sum\nolimits_{h} A^{\ell, h}_{\text{last}, s_k}(\mathbf{h}^{\ell-1}_{s_k} W^{\ell,h}_V) W^{\ell,h}_O.
\end{equation}
where $A^{\ell,h}_{i,j}$ denotes the attention weight assigned by the $h$-th head in the layer $\ell$ from the last position $i$ to subject token $j$.
Here, $\mathbf{a}^\ell_{\text{last}}$ represents the subject-to-last attention contribution at layer $\ell$.
Intuitively, if subject information is critical for prediction, this contribution should have a large norm; otherwise, the norm should be small.

\begin{figure}[t]
\setlength{\abovecaptionskip}{0pt}   
\setlength{\belowcaptionskip}{0pt}
\centering
    \includegraphics[width=0.8\linewidth]{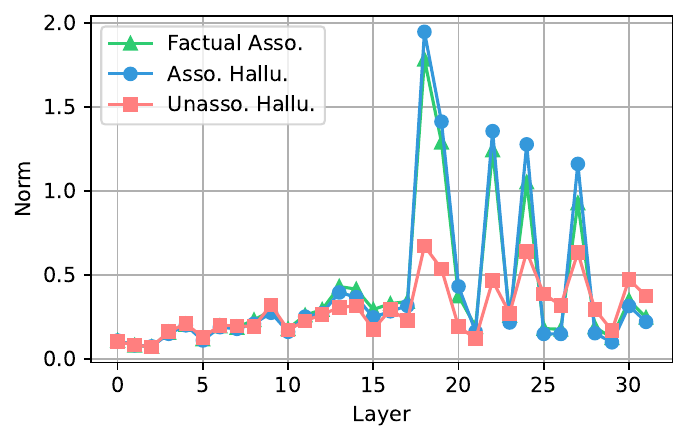}
\caption{
Subject-to-last attention contribution norms across layers in LLaMA-3-8B. Values show the norm of the attention contribution from subject tokens to the last token at each layer.
}
\label{fig:attention-contribution}
\vspace{-3mm}
\end{figure}

Figure~\ref{fig:attention-contribution} shows that in LLaMA-3-8B, both AHs and FAs exhibit large attention-contribution norms in mid-layers, indicating a strong information flow from subject tokens to the target token. In contrast, UHs show consistently lower norms, implying that their predictions rely far less on subject information. \citet{yuksekgonul2023attention} previously argued that high attention flow from subject tokens signals factuality and proposed using attention-based hidden states to detect hallucinations. Our results challenge this view: \textbf{the model propagates subject information just as strongly when generating AHs as when producing correct facts}.

\subsection{Analysis of Last Token Representations}
\label{sec:last-token-representation}

Since both FAs and AHs involve strong information propagation from subject representations, whereas this propagation is weaker for UHs, we hypothesize that this disparity in information flow also leads to distinct geometric properties in the last token representations.
Specifically, the strong subject propagation for FAs and AHs suggests that their last-token representations receive significant subject-specific updates, causing them to \textbf{\textit{diverge}} in representation space. In contrast, because subject information is weakly propagated for UH samples, their last-token representations receive smaller updates and therefore are expected to remain \textbf{\textit{tightly clustered}}, dominated by the shared features of the prompt template.

\begin{figure}[t]
\setlength{\abovecaptionskip}{0pt}   
\setlength{\belowcaptionskip}{0pt}
\centering
    \includegraphics[width=0.8\linewidth]{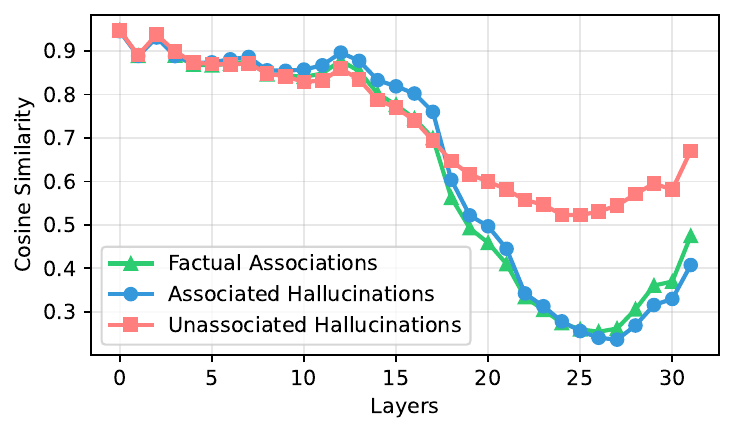}
\caption{
Cosine similarity of target-token hidden states across layers in LLaMA-3-8B. 
}
\label{fig:cosine-similarity-for-the-last-token}
\end{figure}

To verify this, we compute the layer-wise cosine similarity among last-token representations $\mathbf{h}_T^\ell$ within each knowledge category (Figure~\ref{fig:cosine-similarity-for-the-last-token}).
The results indicate that the strength of propagated subject information significantly affects the geometric properties.
In early layers, cosine similarity is high for all categories ($\approx 0.9$).
From the mid-layers onward, where subject representations propagate for FAs and AHs, their last-token representations diverge significantly. By layer 25, the cosine similarity for both FAs and AHs drops to $\approx 0.2$. In contrast, UHs remain moderately clustered, with cosine similarity only declining to $\approx 0.5$.

\begin{figure}[t]
\setlength{\abovecaptionskip}{0pt}   
\setlength{\belowcaptionskip}{0pt}
\centering
    \includegraphics[width=0.8\linewidth]{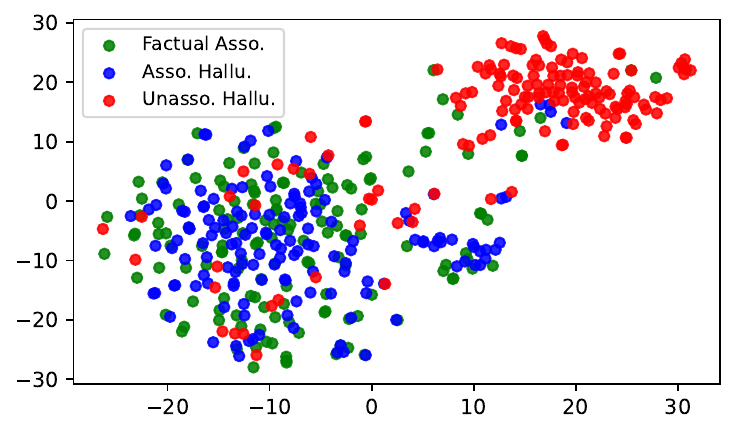}
\caption{
t-SNE visualization of last token's representations at layer 25 of LLaMA-3-8B.
}
\label{fig:last-token-tsne}
\end{figure}

To further illustrate this, we present a t-SNE visualization of the last token's representations at layer 25 of LLaMA-3-8B in Figure \ref{fig:last-token-tsne}. The visualization aligns with our cosine similarity analysis:
hidden representations of UHs form a distinct cluster separated from FAs, whereas AHs substantially overlap with FAs. We quantitatively validate this observation using two cluster separability metrics (the Silhouette coefficient and Davies-Bouldin Index) in Appendix \ref{app:separability-metrics}, which show significantly higher separability for UHs compared to AHs. Additional visualizations can be found in Appendix \ref{app:visualization}.

\begin{figure}[t]
\setlength{\abovecaptionskip}{0pt}   
\setlength{\belowcaptionskip}{0pt}
\centering
    \includegraphics[width=0.85\linewidth]{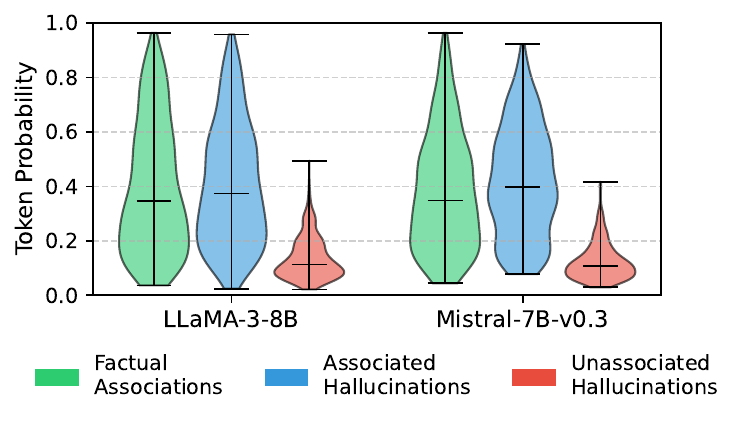}
\caption{
Distribution of last token probabilities. 
}
\label{fig:token-prob-distribution}
\end{figure}

Additionally, we observe that the strength of subject representation propagation also affects the entropy of the output distribution (Figure~\ref{fig:token-prob-distribution}). A strong subject-specific update concentrates the output distribution on specific subject-related entities, resulting in low-entropy outputs for FAs and AHs. Conversely, a weak subject-specific update leads to a wider spread of probability mass covering many possible answers (e.g., assigning probability to multiple possible names for the prompt ``\textit{The name of the father of <subject> is}''), resulting in significantly higher output entropy.

\section{Revisiting Hallucination Detection}
\label{sec:revisiting-detection}
The mechanistic analysis in \S \ref{sec:analyzing-internal-states} reveals that \textit{internal states of LLMs primarily capture \textbf{how the model utilizes its parametric knowledge}, rather than whether the output is truthful}. This suggests that representation-based signals struggle to distinguish associated hallucinations (AHs) from factual associations (FAs) but can more effectively separate unassociated hallucinations (UHs) from FAs.

In this section, we quantify the extent to which existing hallucination detection methods can separate AHs from FAs and compare this performance to detecting UHs.

\paragraph{Experimental Setups}
We revisit the effectiveness of widely-adopted white-box hallucination detection approaches that use internal state probing as well as black-box approaches that rely on scalar features. We evaluate on three settings: 1) \textbf{AH Only} (1,000 FAs and 1,000 AHs for training; 200 of each for testing), 2) \textbf{UH Only} (1,000 FAs and 1,000 UHs for training; 200 of each for testing), and 3) \textbf{Full} (1,000 FAs and 1,000 hallucination samples mixed of AHs and UHs for training; 200 of each for testing). 
For each setting, we repeat data splitting with five random seeds and report mean AUROC and standard deviation.

\noindent\textbf{White-box methods}: 
We extract and normalize internal features and then train a probe.
\begin{itemize}[leftmargin=*,nosep]
    \item \textbf{Subject representations}: last subject token hidden state from three consecutive layers \cite{gottesman2024estimatingknowledgelargelanguage}.
    \item \textbf{Attention flow}: attention weights from the last token to subject tokens across all layers \cite{yuksekgonul2023attention}.
    \item \textbf{Last-token representations}: final token hidden state from the last layer \cite{orgad2024llms}. 
\end{itemize}

\noindent\textbf{Black-box methods}: 
We test two commonly used scalar features, including \textbf{answer token probability} \citep{orgad2024llms} and \textbf{subject popularity} (average monthly Wikipedia page views) \citep{mallen2022not}. As discussed in \S \ref{sec:correlation-with-subject-popularity} and \S \ref{sec:last-token-representation}, these features are also related to whether the model relies on encoded knowledge to produce outputs rather than with truthfulness itself. 

\paragraph{Experimental Results}

\begin{table}[t]
\setlength{\abovecaptionskip}{5pt}   
\setlength{\belowcaptionskip}{0pt}
\centering
\scriptsize
\setlength{\tabcolsep}{3pt}         
\renewcommand{\arraystretch}{0.95}  
\begin{tabular}{@{}l cc cc@{}}      
\toprule
& \multicolumn{2}{c}{\textbf{LLaMA}} & \multicolumn{2}{c}{\textbf{Mistral}} \\
\cmidrule(lr){2-3}\cmidrule(lr){4-5}
\textbf{Methods} & \textbf{AH Only} & \textbf{UH Only} & \textbf{AH Only} & \textbf{UH Only} \\
\midrule
Subject            & $ 0.65 \pm 0.02$ & $ 0.91 \pm 0.01$ & $ 0.57 \pm 0.02$ & $ 0.81 \pm 0.02$ \\
Attention      & $ 0.58 \pm 0.04$ & $ 0.92 \pm 0.02$ & $ 0.58 \pm 0.07$ & $ 0.87 \pm 0.01$ \\
Last Token           & $ \mathbf{0.69 \pm 0.03}$ & $ \mathbf{0.93 \pm 0.01}$ & $ \mathbf{0.63 \pm 0.02}$ & $ \mathbf{0.92 \pm 0.01}$ \\
Probability        & $ 0.49 \pm 0.01$ & $ 0.86 \pm 0.01$ & $ 0.46 \pm 0.00$ & $ 0.89 \pm 0.00$ \\
Subject Pop.       & $ 0.48 \pm 0.01$ & $ 0.87 \pm 0.01$ & $ 0.52 \pm 0.01$ & $ 0.84 \pm 0.01$ \\
\bottomrule
\end{tabular}
\caption{Hallucination detection performance on \textbf{AH Only} and \textbf{UH Only} settings. Detailed statistical significance analysis is provided in Appendix \ref{app:statistical-significance-analysis}.}
\label{tab:probe-perforamce}
\end{table}

\begin{figure}[t]
\setlength{\abovecaptionskip}{0pt}   
\setlength{\belowcaptionskip}{0pt}
\centering
    \includegraphics[width=0.8\linewidth]{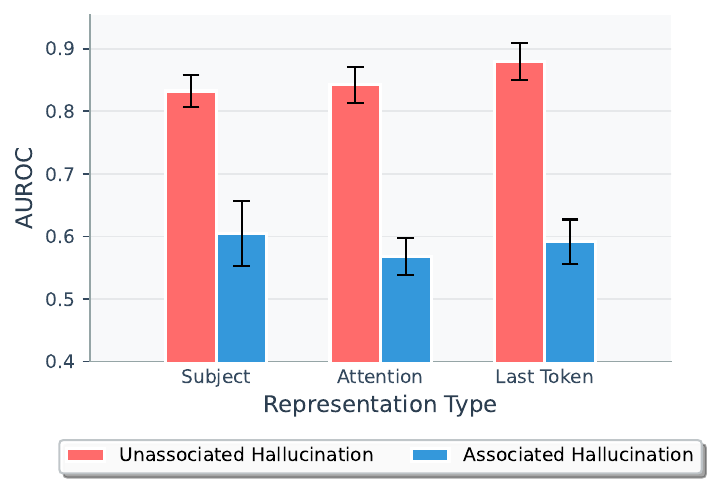}
\caption{
Hallucination detection performance on the \textbf{Full} setting (LLaMA-3-8B).
}
\label{fig:probe-performance}
\end{figure}

Table~\ref{tab:probe-perforamce} shows that hallucination detection methods perform differently in \textbf{AH Only} and \textbf{UH Only} settings. For white-box probes, all approaches effectively distinguish UHs from FAs (AUROC $\approx 0.91$--$0.93$). In contrast, performance drops sharply on the \textbf{AH Only} setting (AUROC $\approx 0.58$--$0.69$). Black-box methods follow the same pattern (AUROC $\approx 0.86$--$0.87$ on \textbf{UH Only} and $\approx 0.48$--$0.49$ on \textbf{AH Only}). Notably, the performance of black-box methods on \textbf{AH Only} is close to random guessing (AUROC $\approx 0.48$--$0.49$), indicating that token uncertainty and subject popularity provide little signal for separating AHs from FAs.
Although white-box representations show slightly better separability than black-box methods, such performance is far from reliable for deployment \cite{hosmer2013applied}. 
Figure~\ref{fig:probe-performance} further highlights this disparity under the \textbf{Full} setting: detection is consistently stronger on UH samples than on AH samples, and adding AHs to the training set significantly dilutes performance on UHs (AUROC $\approx$ 0.9 on \textbf{UH Only} vs. $\approx$ 0.8 on \textbf{Full}).

These results confirm that both internal probes and black-box methods capture whether a model draws on parametric knowledge, not whether its outputs are factually correct. 
Our further analysis in Appendix \ref{app:consistency-detection} reveals that consistency-based detection methods also struggle to distinguish AHs from FAs. 
UHs are easier to detect because they bypass this knowledge, while associated hallucinations are produced through the same recall process as FAs, leaving limited internal cues to distinguish them. 
An analysis of sensitivity-specificity trade-offs in Appendix \ref{app:sensitivity-specificity} further confirms this limitation, showing that while UHs can be detected with high sensitivity and specificity simultaneously, AH detection requires a significant trade-off between them. 
As a result, \textbf{LLMs appear to have limited intrinsic awareness of their own truthfulness, and detection methods relying on these signals risk misclassifying AHs as correct}, fostering harmful overconfidence in model outputs.

\section{Challenges of Refusal Tuning}
\label{sec:refusal-tuning}

\begin{figure}[t]
\setlength{\abovecaptionskip}{0pt}   
\setlength{\belowcaptionskip}{0pt}
\centering
    \includegraphics[width=0.8\linewidth]{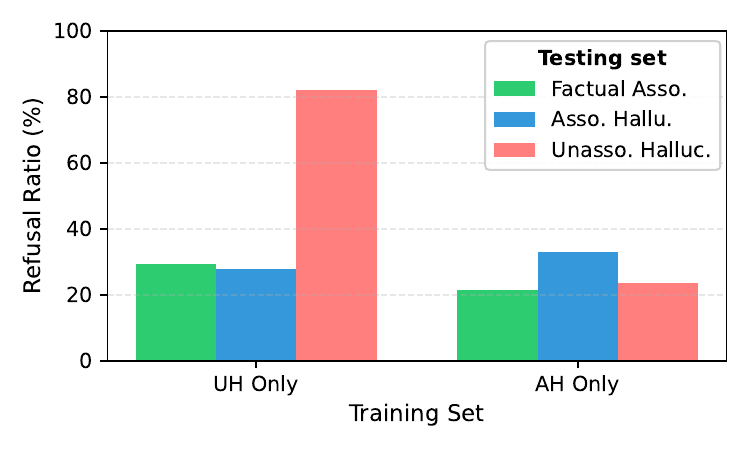}
\caption{
Refusal tuning performance across three types of samples (LLaMA-3-8B). 
}
\label{fig:refusal-tuning-llama}
\end{figure}

A common strategy to mitigate potential hallucinations in the model’s responses is to fine-tune LLMs to refuse answering when they cannot provide a factual response, \textit{e.g.}, \textbf{Refusal Tuning} \cite{zhang2024r}. 
For such refusal capability to generalize, the training data must contain a shared feature pattern across hallucinated outputs, allowing the model to learn and apply it to unseen cases.

Our analysis in the previous sections shows that this prerequisite is not met. Because AH representations are geometrically diverse, it is difficult for models to capture a consistent feature pattern among AH samples, thereby hindering the generalization of refusal responses to unseen AH samples. In contrast, UH representations tend to form distinct cluster, presenting consistent feature patterns that should facilitate easier generalization of the refusal capability.

\paragraph{Experimental Setups}
To verify the hypothesis, we conduct refusal tuning on LLMs under two settings: 1) \textbf{UH Only}, where 1,000 UH samples are paired with 10 refusal templates, and 1,000 FA samples are preserved with their original answers. 2) \textbf{AH Only}, where 1,000 AH samples are paired with refusal templates, with 1,000 FA samples again remain unchanged. We then evaluate both models on 200 samples each of FAs, UHs, and AHs. A response matching any refusal template is counted as a refusal, and we report the \textbf{Refusal Ratio} as the proportion of samples eliciting refusals.
This measures not only whether the model refuses appropriately on UHs and AHs, but also whether it “over-refuses” on FA samples.

\paragraph{Experimental Results}

Figure~\ref{fig:refusal-tuning-llama} shows that training with UHs leads to strong generalization across UHs, with refusal ratios of 82\% for LLaMA. However, this effect does not transfer to AHs, where refusal ratios fall to 28\%, respectively. Moreover, some FA cases are mistakenly refused (29.5\%). These results confirm that UHs share a common activation subspace, supporting generalization within the category, while AHs and FAs lie outside this space. 
By contrast, training with AHs produces poor generalization. On AH test samples, refusal ratio is only 33\%, validating that their subject-specific hidden states prevent consistent refusal learning. Generalization to UHs is also weak (23.5\%), again reflecting the divergence between AH and UH activation spaces.

Overall, these findings indicate that \textbf{the generalizability of refusal tuning is constrained by the heterogeneous nature of hallucinations}. While UH representations are internally consistent enough to support learnable refusal generalization, AH representations are too geometrically diverse for the model to learn a consistent and robust refusal pattern.

\section{Conclusions and Future Work}

In this work, we revisit the widely accepted claim that hallucinations can be detected from a model’s internal states.
Our mechanistic analysis reveals that \textbf{hidden states encode whether models rely on their parametric knowledge rather than truthfulness}.
As a result, detection methods succeed most reliably when outputs are detached from the input but struggle when hallucinations arise from the same knowledge-recall process as correct answers.

These findings lead to three key implications.
First, future evaluations should \textbf{report detection performance separately} for Associated Hallucinations (AHs) and Unassociated Hallucinations (UHs), as they stem from fundamentally different internal processes and require distinct detection strategies. 
Second, relying solely on hidden states is insufficient for reliable hallucination detection. Future research should integrate LLMs with \textbf{external feedback mechanisms}, such as fact-checking modules or retrieval-based verifiers, to assess factuality more robustly. 
Third, future studies should \textbf{prioritize improving AH detection}. Because AHs occur more frequently in widely known or highly popular topics (\S \ref{sec:subject-analysis}), their undetected errors pose greater risks to user trust and the practical reliability of LLMs.

\section*{Limitations}
We identify several limitations of our work.

\paragraph{Focus on Factual Knowledge}
While our analysis identifies challenging cases of hallucination detection methods, our study is primarily limited to factual completion prompts. It does not extend to long-form or open-ended text generation tasks \cite{wei2024long, Min2023FActScoreFA, Huang2024FactAlignLF}. Future work should broaden this investigation to these tasks in order to draw more comprehensive conclusions.

\paragraph{Lack of Analysis on Prompt-based Hallucination Detection Approaches}
Our analysis focuses on white-box hallucination detection methods based on internal states and two black-box approaches based on external features. We do not include verbalization-based strategies \cite{DBLP:journals/tmlr/LinHE22, DBLP:conf/emnlp/TianMZSRYFM23, DBLP:conf/iclr/XiongHLLFHH24, DBLP:conf/nips/0004CQN024, DBLP:conf/acl/NiBGC24,DBLP:conf/naacl/ZhaoY0XMWCRY24}, such as prompting the model to report or justify its confidence explicitly, which constitute a different line of approach. Exploring such approaches may offer complementary insights into how models internally represent and express uncertainty.

\paragraph{Applicability to Black-box LLMs or Large Reasoning Models}
Our study is limited to open-source LLMs. Conducting mechanistic analyses on commercial black-box LLMs is not permitted due to access restrictions. Future work could explore alternative evaluation protocols or collaboration frameworks that enable partial interpretability analyses on such systems. In addition, recent studies \cite{lrm1,zhang2025selfawarenesslargereasoningmodels} have begun examining the internal states of large reasoning models for hallucination detection, suggesting a promising direction for extending our methodology to models with multi-step reasoning capabilities.

\section*{Ethical Considerations}

This work analyzes the internal mechanisms of large language models using data constructed from Wikidata \cite{DBLP:journals/cacm/VrandecicK14}, which is released under the Creative Commons CC0 1.0 Universal license, allowing unrestricted use and redistribution of its data. All data are derived from publicly available resources, and no private or sensitive information about individuals is included. We used LLM tools for language polishing.

\section*{Acknowledgments}
This research was supported by the Singapore Ministry of Education (MOE)
Academic Research Fund (AcRF) Tier 1 grant (Proposal ID: 24-SIS-SMU-002) and the National Research Foundation Singapore under the AI Singapore Programme (AISG Award No: AISG3-RPGV-2025-016).

\bibliography{custom}

\appendix

\section*{Appendix}

\section{Datasets and Implementations}
\label{app:data}

\subsection{Selected Relations and Prompt Templates}
\label{sec:relation-and-prompt-template}
We employed a set of criteria to select relations from Wikidata in order to construct our dataset. Our criteria largely follow the framework proposed by \citet{DBLP:journals/corr/abs-2503-15299}.
Specifically, we require that each factual query in the dataset be unambiguous: given a subject–relation pair, the object should be unique and easy verifiable.
The criteria are as follows: 
\begin{itemize}[leftmargin=*]
    \item \textbf{Avoid granularity ambiguity}. We exclude relations whose answers can vary in their level of detail. For example, in \textit{location} queries, the response could be expressed as a city, state, or country, making it ill-defined \cite{DBLP:conf/acl/YonaAG24}.
    \item \textbf{Avoid surface-level guessing.} We exclude relations whose correct answers can often be inferred from shallow patterns. For instance, \textit{country of citizenship} can frequently be guessed from shallow lexical patterns, rather then reflecting actual memorization \cite{DBLP:conf/acl/MallenAZDKH23}.
\end{itemize}

Following these criteria, \citet{DBLP:journals/corr/abs-2503-15299} narrowed the 24 relations introduced by \citet{DBLP:conf/emnlp/SciavolinoZLC21} down to four. However, we observe that their filtering primarily addresses ambiguity at the relation and object levels, but does not consider ambiguity at the subject level. In practice, some relations involve subjects that are inherently ambiguous. For example, the relation \textit{record label} can be problematic because many songs share identical names, leading to unclear subject–object mappings.

To mitigate such cases, we apply an additional subject-level filtering step and restrict our dataset to relations where the subject is a person, thereby reducing ambiguity. In addition, we manually include certain relations to strengthen the dataset. Concretely, we use the following four relations: P22 (\textit{father}), P25 (\textit{mother}), P26 (\textit{spouse}), and P569 (\textit{date of birth}). We show the list of the templates used to create our dataset in Table~\ref{tab:list-of-promopts}.

\begin{table}[t]
\footnotesize
\renewcommand{\arraystretch}{1.2}
\setlength{\tabcolsep}{2pt}
    \centering
    \begin{tabular}{lr}
\toprule
\textbf{Relation } & \textbf{Prompt Template} \\\midrule
father  & The name of the father of \texttt{[subject]} is \\
mother  & The name of the mother of \texttt{[subject]} is \\
spouse  & The name of the spouse of \texttt{[subject]} is \\
date of birth & The birth date of \texttt{[subject]} is \\
\bottomrule
 \end{tabular}
    \caption{Relations and prompt templates for querying factual knowledge of models. \texttt{[subject]} is a placeholder replaced with subject entities.}
    \label{tab:list-of-promopts}
\end{table}

\begin{figure*}[t]
\centering
\begin{tabular}{|p{0.9\linewidth}|}
\hline
\textbf{LLM Judge Prompt} \\ \hline
I will give you a factual query (e.g., ``The name of the father of <subj>''), a gold answer to the factual query, and a proposed answer generated by an LLM. You need to compare the proposed answer to the gold answer and assign it one of the possible grades using the steps below. \\[0.8ex]

\textbf{Possible grades are:} \\
A: CORRECT \\
B: INCORRECT \\
C: WRONG GOLD \\
D: ERROR \\[0.8ex]

Spelling errors, synonyms, abbreviations, or hedging expressions (e.g., ``it is possible that'') should not alter the grade if the person referred to in the proposed answer matches the gold answer. \\[0.8ex]

\textbf{Steps:} \\
Step 1: If the gold answer does not correspond to an answer for the question, output ``C'' and finish. Otherwise, proceed to Step 2. \\
Step 2: Extract all predicted entities from the proposed answer. Proceed to Step 3. \\
Step 3: If each predicted entity refers to the answer mentioned in the gold answer, output ``A'' and finish. Otherwise, proceed to Step 4. \\
Step 4: If the predicted entity does not refer to the gold answer, output ``B'' and finish. Otherwise, proceed to Step 5. \\
Step 5: Double-check whether the proposed answer refers to a different answer from the gold answer. If it does, output ``B.'' Otherwise, output ``D'' and finish. \\[0.8ex]

\textbf{Input format:} \\
Question: \{question\} \\
Gold answer: \{gold\_answer\} \\
Proposed answer: \{proposed\_answer\} \\[0.8ex]

\textbf{Instruction:} Output your reasoning steps. After that, conclude your response with ``Output:'' followed by the letter (A, B, C, or D). Do not provide any further explanation. \\
\hline
\end{tabular}
\caption{LLM Judge prompt used for evaluation.}
\label{fig:llm_judge_prompt}
\end{figure*}

\begin{figure*}[t]
\setlength{\abovecaptionskip}{5pt}   
\setlength{\belowcaptionskip}{0pt}
  \centering
  \begin{subfigure}[t]{0.3\linewidth}
    \includegraphics[width=\linewidth]{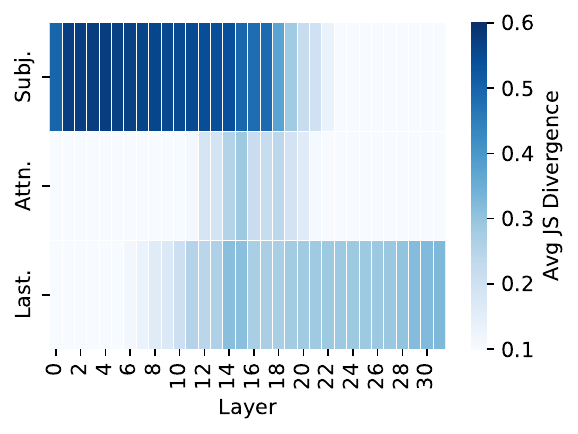}
    \caption{Factual Associations}
    \label{fig:interventions-mistral-correct}
  \end{subfigure} \hfill
  \begin{subfigure}[t]{0.3\linewidth}
    \includegraphics[width=\linewidth]{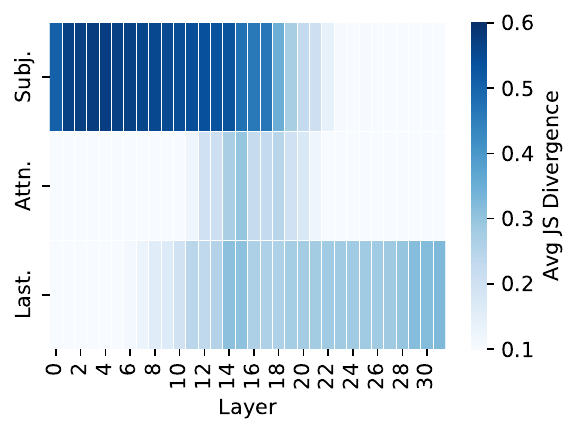}
    \caption{Associated Hallucinations}
    \label{fig:interventions-mistral-ah}
  \end{subfigure} \hfill
  \begin{subfigure}[t]{0.3\linewidth}
    \includegraphics[width=\linewidth]{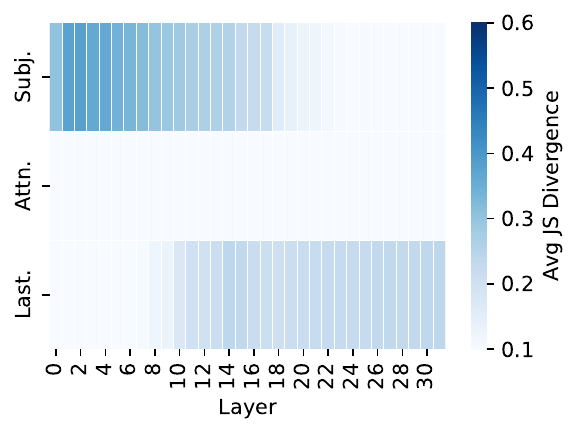}
    \caption{Unassociated Hallucinations}
    \label{fig:interventions-mistral-uh}
  \end{subfigure} 
  \caption{Effect of interventions across layers of Mistral-7B-v0.3. The heatmap shows JS divergence between the  output distribution before and after intervention. Darker color indicates that the intervened hidden states are more causally influential on the model’s predictions.
Top row: patching representations of subject tokens. 
Middle row: blocking attention flow from subject to the last token.  
Bottom row: patching representations of the last token.}
  \label{fig:interventions-mistral}
\end{figure*}

\subsection{Implementation Details}\label{app:implement}

\paragraph{Checkpoints and GPU resources.}

All the checkpoints used in our experiments are provided by the Hugging Face Transformers library \cite{DBLP:journals/corr/abs-1910-03771}. Specifically, we use the checkpoint ``meta-llama/Meta-Llama-3-8B''\footnote{\url{https://huggingface.co/meta-llama/Meta-Llama-3-8B}} and ``mistralai/Mistral-7B-v0.3''\footnote{\url{https://huggingface.co/mistralai/Mistral-7B-v0.3}} for the experiments of response generation (\S \ref{sec:data}), hidden-state analysis (\S \ref{sec:analyzing-internal-states}) and accessing the performance of hallucination detection methods (\S \ref{sec:revisiting-detection}).
For refusal tuning (\S\ref{sec:refusal-tuning}), we use checkpoints provided by the Unsloth framework \cite{unsloth}, namely ``unsloth/llama-3-8b''\footnote{\url{https://huggingface.co/unsloth/llama-3-8b}} and ``unsloth/mistral-7b-v0.3''\footnote{\url{https://huggingface.co/unsloth/mistral-7b-v0.3}}, which enable more efficient fine-tuning.
All experiments are conducted on 4 NVIDIA L40S GPUs.

\begin{figure}[t]
\setlength{\abovecaptionskip}{5pt}   
\setlength{\belowcaptionskip}{0pt}
\centering
    \includegraphics[width=0.8\linewidth]{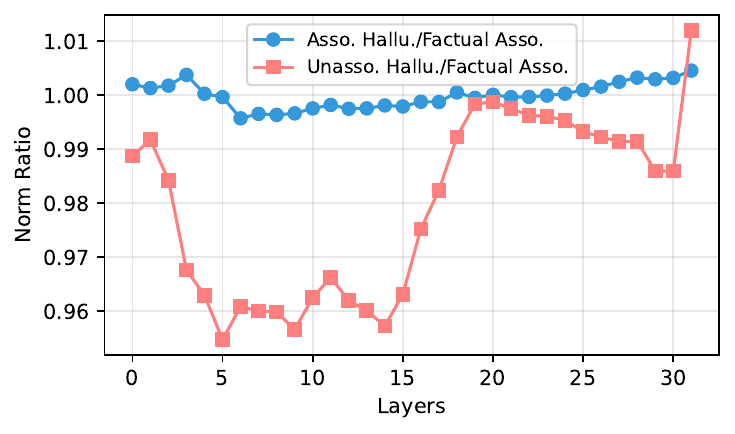}
\caption{
Norm ratio curves of subject representations in Mistral-7B-v0.3, comparing AHs and UHs against FAs as the baseline. At earlier layers, the norm of UH samples is significantly lower than that of AH samples.
}
\label{fig:subject_resp_norm_ratio-mistral}
\end{figure}

\begin{figure}[t]
\setlength{\abovecaptionskip}{5pt}   
\setlength{\belowcaptionskip}{0pt}
\centering
    \includegraphics[width=0.8\linewidth]{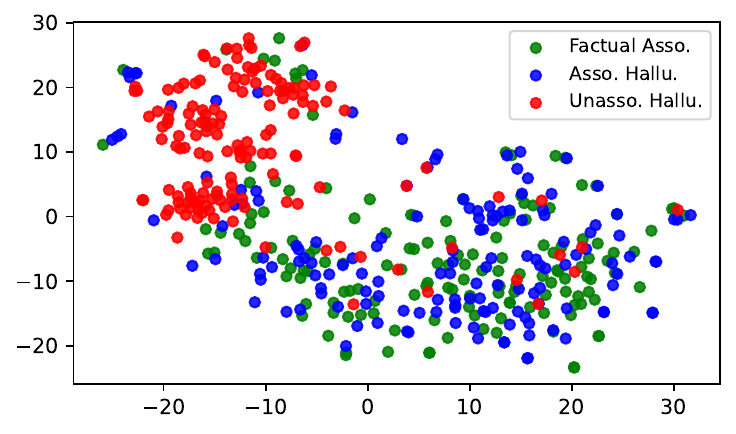}
\caption{
t-SNE visualization of last token's representations at layer 25 of Mistral-7B-v0.3.
}
\label{fig:last-token-tsne-mistral}
\end{figure}

\begin{figure}[t]
\setlength{\abovecaptionskip}{5pt}   
\setlength{\belowcaptionskip}{0pt}
\centering
    \includegraphics[width=0.8\linewidth]{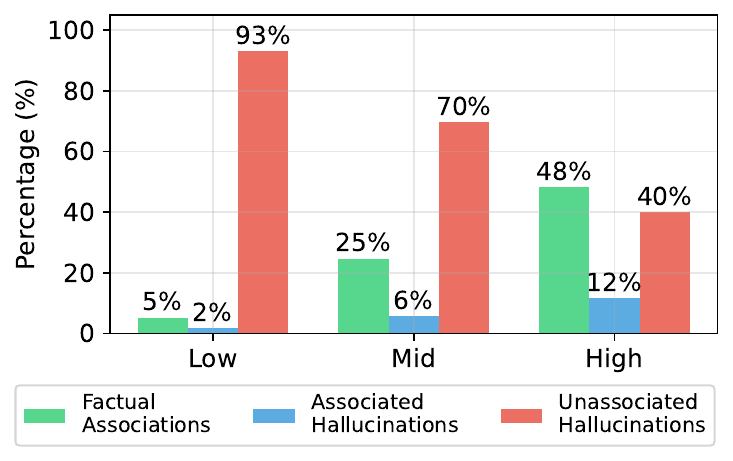}
\caption{
Sample distribution across different subject popularity (low, mid, high) in Mistral-7B-v0.3, measured by monthly Wikipedia page views. }
\label{fig:error-distribution-mistral}
\end{figure}

\paragraph{Decoding algorithm.}
We employ greedy decoding ($\text{temperature} = 0$) for response generation, with models run in BF16 precision.  

\paragraph{PEFT settings for refusal tuning.}
For refusal tuning, we fine-tune with both models using QLoRA \cite{DBLP:conf/nips/DettmersPHZ23}, implemented with the Unsloth framework \cite{unsloth}, with rank $r=8$, and $\alpha=8$. QLoRA adapters are applied to all attention and MLP modules, and each model is fine-tuned for one epoch.

\subsection{Labeling Factual Correctness}
\label{app:label}
Here, we outline the process of determine Factual Correctness of a model prediction.
We construct correctness labels through a two-stage process. First, we use spaCy\footnote{\url{https://spacy.io/}} Named Entity Recognizer to extract the target entity from the model’s output. If it matches the ground truth, the answer is marked correct. Otherwise, or if extraction fails, we rely on Qwen2.5-14B-Instruct \cite{qwen2.5} as a judge to compare the predicted answer with the ground truth. Following \citet{DBLP:journals/corr/abs-2503-15299}, we design the evaluation prompt, which is shown in Figure~\ref{fig:llm_judge_prompt}.

\begin{figure}[t]
\setlength{\abovecaptionskip}{5pt}   
\setlength{\belowcaptionskip}{0pt}
\centering
    \includegraphics[width=0.8\linewidth]{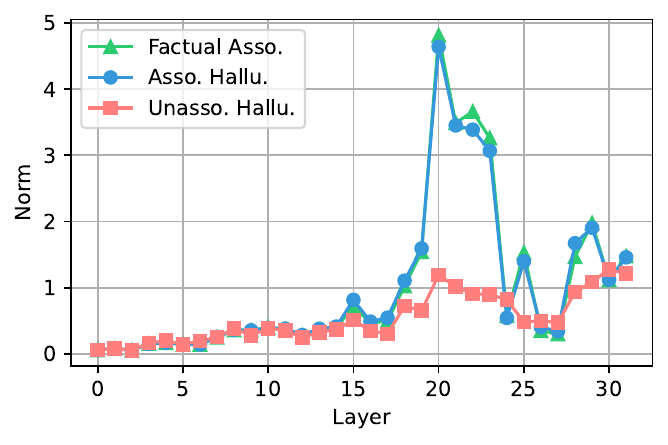}
\caption{
Subject-to-last attention contribution norms across layers in Mistral-7B-v0.3. Values show the norm of the attention contribution from subject tokens to the last token at each layer.
}
\label{fig:attention-contribution-mistral}
\end{figure}

\begin{figure}[t]
\setlength{\abovecaptionskip}{5pt}   
\setlength{\belowcaptionskip}{0pt}
\centering
    \includegraphics[width=0.8\linewidth]{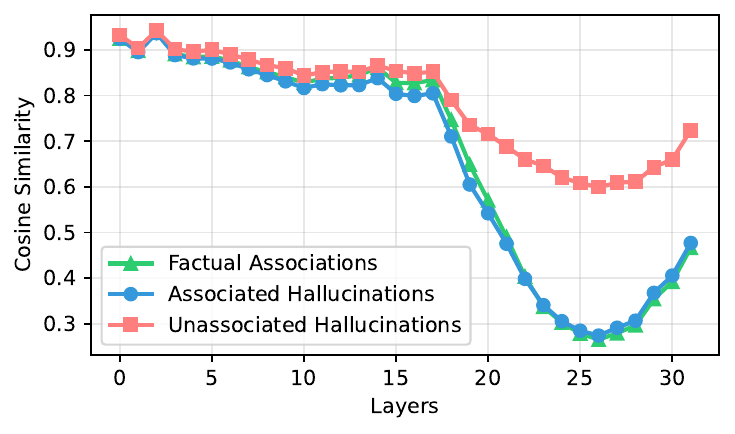}
\caption{
Cosine similarity of target-token hidden states across layers in Mistral-7B-v0.3. From mid-layers onward, FAs and AHs diverge sharply as subject information propagates, while UHs remain more clustered, confirming weaker subject-dependent updates.
}
\label{fig:cosine-similarity-for-the-last-token-mistral}
\end{figure}

\begin{figure}[t]
\setlength{\abovecaptionskip}{5pt}   
\setlength{\belowcaptionskip}{0pt}
\centering
    \includegraphics[width=0.8\linewidth]{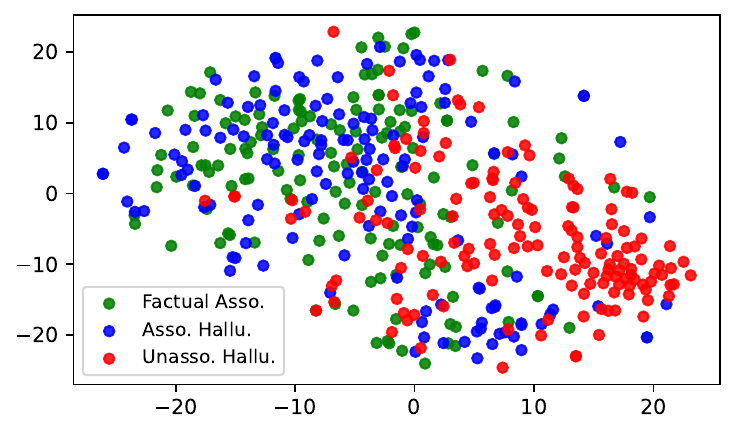}
\caption{
t-SNE visualization of subject tokens' representations at layer 11 of LLaMA-3-8B.
}
\label{fig:subject-tsne-llama}
\end{figure}

\begin{figure}[t]
\setlength{\abovecaptionskip}{5pt}   
\setlength{\belowcaptionskip}{0pt}
\centering
    \includegraphics[width=0.8\linewidth]{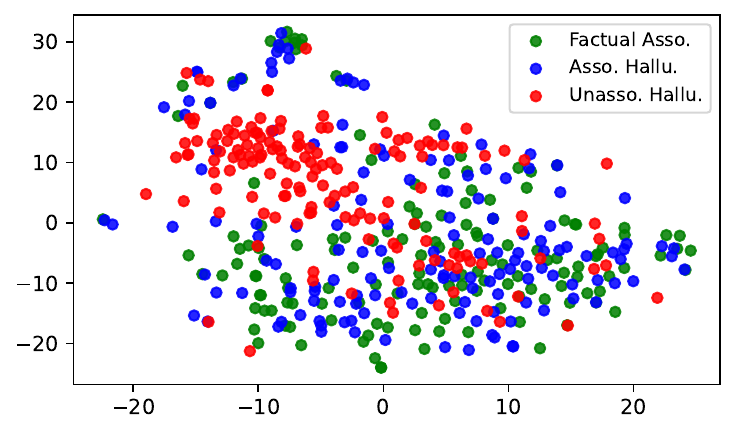}
\caption{
t-SNE visualization of subject tokens' representations at layer 11 of Mistral-7B-v0.3.
}
\label{fig:subject-tsne-mistral}
\end{figure}

\begin{figure}[t]
\setlength{\abovecaptionskip}{5pt}   
\setlength{\belowcaptionskip}{0pt}
\centering
    \includegraphics[width=0.8\linewidth]{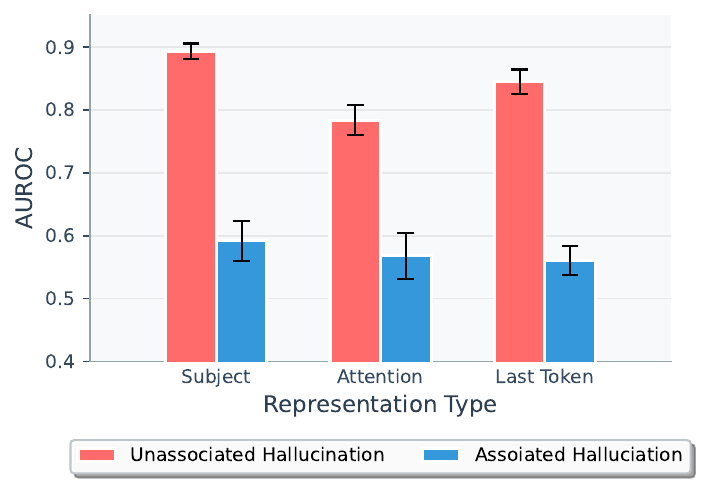}
\caption{
Hallucination detection performance on the
\textbf{Full} setting (Mistral-7B-v0.3).
}
\label{fig:probe-performance-mistral}
\end{figure}

\begin{figure}[t]
\setlength{\abovecaptionskip}{5pt}   
\setlength{\belowcaptionskip}{0pt}
\centering
    \includegraphics[width=0.8\linewidth]{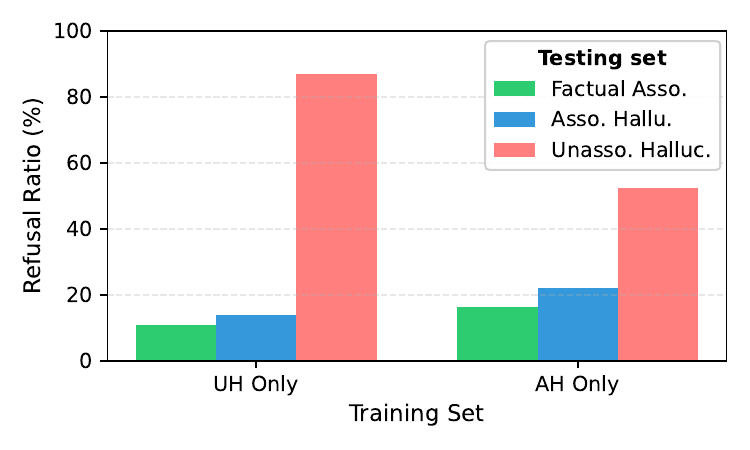}
\caption{
Refusal tuning performance across three
types of samples (Mistral-7B-v0.3).
}
\label{fig:refusal-tuning-mistral}
\end{figure}

\begin{figure}[t]
\setlength{\abovecaptionskip}{5pt}   
\setlength{\belowcaptionskip}{0pt}
\centering
    \includegraphics[width=0.8\linewidth]{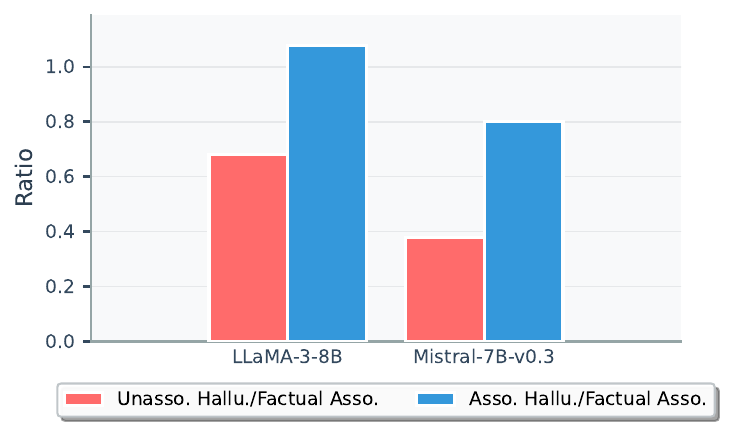}
\caption{
Comparison of subspace overlap ratios.
}
\label{fig:subspace-overlap}
\end{figure}

\section{Calculation of Subject Representation Norms}
\label{sec:appendix_norm_details}

To investigate the geometric properties of subject representations, we compute the average $L_2$ norm of the hidden states across different knowledge categories. For a single sample with subject tokens $s_1, \ldots, s_n$, we first compute the average norm across all subject tokens at layer $\ell$:

\begin{equation}
\bar{N}^{(\ell)}_{\text{sample}} = \frac{1}{n} \sum_{i=1}^{n} \left\| \mathbf{h}^{(\ell)}_{s_i} \right\|_2
\end{equation}

Then, for a given category (e.g., FAs, UHs, or AHs) containing $K$ samples, we aggregate these values to obtain the per-category average norm:
\begin{equation}
\bar{N}^{(\ell)}_{\text{category}} = \frac{1}{K} \sum_{j=1}^{K} \bar{N}^{(\ell)}_{\text{sample}, j}
\end{equation}
where $\bar{N}^{(\ell)}{\text{sample}, j}$ is the average norm for the $j$-th sample in the category. By repeating this process for each layer $\ell \in \{1, \ldots, L\}$, we obtain a layer-wise norm profile. Finally, to facilitate direct comparison, we report the norm ratio of hallucinations relative to factual associations:

\begin{align}
\text{Ratio}^{(\ell)}_{\text{AH/FA}} &= \frac{\bar{N}^{(\ell)}_{\text{AH}}}{\bar{N}^{(\ell)}_{\text{FA}}} \\[5pt]
\text{Ratio}^{(\ell)}_{\text{UH/FA}} &= \frac{\bar{N}^{(\ell)}_{\text{UH}}}{\bar{N}^{(\ell)}_{\text{FA}}}
\end{align}

\section{Parallel Experiments on Mistral}
\label{app:mistral}

This section is for documenting parallel experiments conducted on the Mistral-7B-v0.3 model under the same settings as described in the main text (Figures~\ref{fig:interventions-mistral}--\ref{fig:refusal-tuning-mistral}).

The results from Mistral exhibit similar patterns to those observed in LLaMA, as described before. Specifically, we find consistent patterns in the model’s internal computations, hidden-state behaviors, and the performance of hallucination detection and refusal tuning experiments.

\section{Further Analysis on Internal States}

\subsection{Analysis of Parametric Subspace Overlap for Subject Representation}
\label{app:subject-representation-subspace-overlap}

We investigate why early layers encode subject representations differently across knowledge types by examining how inputs interact with the parametric knowledge stored in MLP modules. Inspired by \citet{kang2023deep}, the output norm of an MLP layer depends on how well its input aligns with the subspace spanned by the weight matrix: \textit{poorly aligned inputs yield smaller output norms}. 

For each MLP layer $\ell$, we analyze the down-projection weight matrix $W_\text{down}^\ell$ and its input $x^\ell$. Given the input $x_s^\ell$ corresponding to the subject tokens, we compute its overlap ratio with the top singular subspace $V_\text{top}$ of $W_\text{down}^\ell$:
\begin{equation}\small
    r(x_s^\ell)=\frac{\left\lVert {x_s^\ell}^\top V_{\text{top}} V_{\text{top}}^\top \right\rVert^2}{\left\lVert x_s^\ell \right\rVert^2}.
\end{equation}
A higher overlap ratio $r(x_s^\ell)$ indicates stronger alignment to the subspace spanned by $W_\text{down}^\ell$, leading to larger output norms. 

To highlight relative deviations from the factual baseline (FA), we report the relative ratios between AH/FA and UH/FA. 
Focusing on the layer with the largest UH norm shift, Figure~\ref{fig:subspace-overlap} shows that UHs have significantly lower $r(x_s^\ell)$ than AHs in both LLaMA and Mistral. This reveals that early-layer parametric weights are more aligned with FA and AH subject representations than with UH subjects, producing higher norms for the former ones. These results also suggest that the model has sufficiently learned representations for FA and AH subjects during pretraining but not for UH subjects.

\subsection{Quantitative Analysis of Cluster Separability}
\label{app:separability-metrics}

To quantitatively validate the visual separability observed in the t-SNE visualizations (Figures~\ref{fig:last-token-tsne} and~\ref{fig:last-token-tsne-mistral}), we compute two cluster separability metrics on the hidden representations: the Silhouette coefficient and the Davies-Bouldin Index (DBI). The Silhouette coefficient ranges from $-1$ to $1$, where higher values indicate better-defined clusters. The DBI measures the average similarity ratio of each cluster with its most similar cluster, where lower values indicate better separation.

\begin{table}[t]
\centering
\footnotesize
\setlength{\tabcolsep}{3pt}
\begin{tabular}{lcccc}
\toprule
& \multicolumn{2}{c}{\textbf{UH Only}} & \multicolumn{2}{c}{\textbf{AH Only}} \\
\cmidrule(lr){2-3} \cmidrule(lr){4-5}
\textbf{Model} & \textbf{Sil.} $\uparrow$ & \textbf{DBI} $\downarrow$ & \textbf{Sil.} $\uparrow$ & \textbf{DBI} $\downarrow$ \\
\midrule
LLaMA   & 0.397 & 0.932  & 0.002  & 14.22 \\
Mistral & 0.327 & 1.134  & -0.001 & 25.25 \\
\bottomrule
\end{tabular}
\caption{Quantitative cluster separability metrics. Higher Silhouette (Sil., $\uparrow$) and lower Davies--Bouldin Index (DBI, $\downarrow$) indicate better separation.}
\label{tab:separability-metrics}
\end{table}

The results shown in Table \ref{tab:separability-metrics} quantitatively confirm our observations from the t-SNE visualizations. When comparing UHs and FAs, both models exhibit high Silhouette coefficients (0.33--0.40) and low DBI values (0.93--1.13), indicating clear cluster separation between UHs and FAs. In contrast, when comparing AHs and FAs, the Silhouette coefficients are near zero (0.0022 and $-0.0011$) and DBI values are substantially higher (14.22 and 25.25), indicating strong overlap between AHs and FAs. These metrics support our claim that FAs and UHs form well-separated clusters in the representation space, while AHs strongly overlap with FAs, consistent with our hypothesis that AHs rely on similar knowledge-recall processes as FAs.

\subsection{More Visualization on Hidden States}
\label{app:visualization}

In this section, we provide t-SNE visualization of subject tokens' hidden states in Figure \ref{fig:subject-tsne-llama} and Figure \ref{fig:subject-tsne-mistral}.

Compared to the last-token representations, the t-SNE visualization of subject-token hidden states shows that unassociated hallucinations (UHs) are moderately separated from factual and associated samples, they exhibit marginally lower separability than last-token representations.
This observation aligns with the results in \S\ref{sec:revisiting-detection}, where the hallucination detection performance using last-token hidden states outperforms that based on subject-token representations.

\section{Further Analysis on Hallucination Detection}
\subsection{Statistical Significance Analysis}
\label{app:statistical-significance-analysis}

We report the hallucination detection performance (\S\ref{sec:revisiting-detection}) with 95\% confidence intervals across five random seeds in Table~\ref{tab:probe-performance-ci}.
Across all settings relevant to our claims, AUROC on UHs is significantly higher than on AHs (all $p<0.01$), supporting our finding that current detection methods are substantially more effective for UHs than AH. 

\begin{table}[t]
\centering
\footnotesize
\setlength{\tabcolsep}{4pt}
\begin{tabular}{llcc}
\toprule
\textbf{Model} & \textbf{Method} & \textbf{UH Only} & \textbf{AH Only} \\
\midrule
\multirow{5}{*}{Mistral} 
& Prob.      & .894 (.889--.899) & .457 (.453--.461) \\
& Subj. Pop. & .812 (.787--.839) & .570 (.551--.590) \\
& Subject    & .893 (.878--.909) & .592 (.552--.631) \\
& Attention  & .872 (.864--.881) & .584 (.500--.668) \\
& Last Token & .923 (.913--.934) & .629 (.610--.648) \\
\midrule
\multirow{5}{*}{LLaMA} 
& Prob.      & .860 (.853--.867) & .488 (.480--.495) \\
& Subj. Pop. & .869 (.856--.883) & .479 (.467--.491) \\
& Subject    & .913 (.897--.930) & .648 (.626--.669) \\
& Attention  & .920 (.889--.950) & .579 (.525--.634) \\
& Last Token & .935 (.923--.947) & .691 (.660--.722) \\
\bottomrule
\end{tabular}
\caption{Hallucination detection performance with 95\% confidence intervals across five random seeds.}
\label{tab:probe-performance-ci}
\end{table}

\subsection{Semantic Consistency-Based Detection}
\label{app:consistency-detection}

We sample 200 instances per category and generate 5 responses per prompt with temperature = 0.7. We compute a consistency score for each sample and evaluate hallucination detection performance using these scores, measured by AUROC. Table~\ref{tab:consistency-results} shows the results.

\begin{table}[t]
\centering
\small
\begin{tabular}{lcc}
\toprule
\textbf{Model} & \textbf{UH Only} & \textbf{AH Only} \\
\midrule
LLaMA   & 0.8374 & 0.6183 \\
Mistral & 0.8437 & 0.6159 \\
\bottomrule
\end{tabular}
\caption{Semantic consistency-based hallucination detection performance.}
\label{tab:consistency-results}
\end{table}

The results show that the performance of this consistency-based method aligns with our findings for both the black-box and white-box methods discussed in \S \ref{sec:revisiting-detection}: it can effectively distinguish UHs from FAs (AUROC $\approx$ 0.84) but struggles to separate AHs from FAs (AUROC $\approx$ 0.62). This behavioral evidence supports our mechanistic analysis: since AHs rely on similar knowledge-recall processes as FAs, they tend to produce similarly consistent outputs across runs, whereas UHs exhibit higher variance due to weaker reliance on subject representations.

\begin{table}[t]
\centering
\small
\setlength{\tabcolsep}{3pt} 
\begin{tabular}{ccccc}
\toprule
\multirow{2}{*}{\textbf{Threshold}} & \multicolumn{2}{c}{\textbf{UH Only}} & \multicolumn{2}{c}{\textbf{AH Only}}\\
\cmidrule(lr){2-3} \cmidrule(lr){4-5}
& \textbf{Sensitivity} & \textbf{Specificity} & \textbf{Sensitivity} & \textbf{Specificity} \\
\midrule
0.1 & 0.900 & 0.833 & 0.697 & 0.583\\
0.3 & 0.879 & 0.888 & 0.644 & 0.633\\
0.5 & 0.867 & 0.925 & 0.601 & 0.679\\
0.7 & 0.867 & 0.942 & 0.534 & 0.725\\
0.9 & 0.838 & 0.963 & 0.433 & 0.771\\
\bottomrule
\end{tabular}
\caption{Sensitivity-specificity trade-offs at different thresholds.}
\label{tab:tradeoff}
\end{table}

\subsection{Sensitivity-Specificity Trade-offs}
\label{app:sensitivity-specificity}

To further examine the detection performance beyond AUROC, we analyze the trade-offs between sensitivity and specificity at various classification thresholds. We report results for detectors trained on LLaMA's last-token representations. 
Tables~\ref{tab:tradeoff} present the results.

For UH Only, we observe that detection performance is remarkably robust across the threshold range. Even as the threshold increases from 0.1 to 0.9, the detector maintains a high balance of metrics, with sensitivity remaining above 0.83 and specificity exceeding 0.96 at the highest setting. This stability indicates a clear and consistent separation in the representation space.
In contrast, for AH Only, the performance is highly sensitive to the specific threshold. As the threshold moves from 0.1 to 0.9, sensitivity drops significantly from 0.697 to 0.433, while specificity only improves from 0.583 to 0.771. This trade-off suggests that establishing a stable decision boundary between AHs and FAs is challenging, confirming that their internal representations strongly overlap.

\end{document}